\documentclass[sigconf]{acmart}

\usepackage{amsthm}
\usepackage{amsmath}
\usepackage{color}
\usepackage{enumitem}
\usepackage{graphicx}
\usepackage{multirow}
\usepackage{pifont}
\usepackage{tabularx}
\usepackage{subfigure}
\usepackage{xcolor}
\usepackage[ruled,vlined,linesnumbered]{algorithm2e}

\AtBeginDocument{%
  }

\setcopyright{acmlicensed}
\copyrightyear{2026}
\acmYear{2026}
\acmDOI{XXXXXXX.XXXXXXX}

\acmConference[The ACM Web Conference '26]{The Web Conference 2026}{April 13--17,
  2026}{Dubai, United Arab Emirates}
\acmISBN{978-1-4503-XXXX-X/2018/06}

\acmSubmissionID{1219}

\begin{document}
\title{FairGU: Fairness-aware Graph Unlearning in Social Networks}

\author{Renqiang Luo}
\affiliation{%
    \institution{Jilin University}
    \city{Changchun}
    \country{China}
}
\email{lrenqiang@outlook.com}

\author{Yongshuai Yang}
\affiliation{%
    \institution{Zhejiang Gongshang University}
    \city{Hangzhou}
    \country{China}
    }
\email{frost.yang@outlook.com}

\author{Huafei Huang}
\affiliation{%
    \institution{Adelaide University}
    \city{Adelaide}
    \country{Australia}
    }
\email{hhuafei@outlook.com}

\author{Qing Qing}
\affiliation{%
    \institution{Jilin University}
    \city{Changchun}
    \country{China}
}
\email{qingqingbai@outlook.com}

\author{Mingliang Hou}
\affiliation{%
    \institution{Jinan University \& TAL Education Group}
    \city{Guangzhou}
    \country{China}
}
\email{teemohold@outlook.com}

\author{Ziqi Xu}
\affiliation{%
    \institution{RMIT University}
    \city{Melbourne}
    \country{Australia}
}
\email{ziqi.xu@rmit.edu.au}

\author{Yi Yu}
\affiliation{%
    \institution{Nanyang Technological University}
    \country{Singapore}
}
\email{yuyi0010@e.ntu.edu.sg}

\author{Jingjing Zhou}
\authornote{Corresponding author.}
\affiliation{%
    \institution{Zhejiang Gongshang University}
    \city{Hangzhou}
    \country{China}
    }
\email{zhoujingjing@zjgsu.edu.cn}

\author{Feng Xia}
\affiliation{%
  \institution{RMIT University}
  \city{Melbourne}
  \country{Australia}
}
\email{f.xia@ieee.org}

\renewcommand{\shortauthors}{Luo et al.}

\begin{abstract}
Graph unlearning has emerged as a critical mechanism for supporting sustainable and privacy-preserving social networks, enabling models to remove the influence of deleted nodes and thereby better safeguard user information. 
However, we observe that existing graph unlearning techniques insufficiently protect sensitive attributes, often leading to degraded algorithmic fairness compared with traditional graph learning methods. 
To address this gap, we introduce FairGU, a fairness-aware graph unlearning framework designed to preserve both utility and fairness during the unlearning process. 
FairGU integrates a dedicated fairness-aware module with effective data protection strategies, ensuring that sensitive attributes are neither inadvertently amplified nor structurally exposed when nodes are removed. 
Through extensive experiments on multiple real-world datasets, we demonstrate that FairGU consistently outperforms state-of-the-art graph unlearning methods and fairness-enhanced graph learning baselines in terms of both accuracy and fairness metrics. 
Our findings highlight a previously overlooked risk in current unlearning practices and establish FairGU as a robust and equitable solution for the next generation of socially sustainable networked systems.
The codes are available at ~\url{https://github.com/LuoRenqiang/FairGU}.
\end{abstract}




\maketitle

\section{Introduction}
\par As web applications continue to evolve into complex, interconnected ecosystems, graph-structured data has become the foundation of many large-scale online services~\cite{zhang2024logical}, including social media~\cite{vombatkere2024tiktok,DuLC0GC025} and e-commerce~\cite{he2024ai}. 
In these settings, removing some privacy-related information from trained graph models is urgent~\cite{li2025community}.
Graph unlearning has emerged as a critical capability for protecting user autonomy and supporting ethical data governance~\cite{fan2025opengu}. 
By enabling a model to remove a user’s data and its downstream influence, graph unlearning directly supports core values of an inclusive and sustainable web~\cite{chen2025graph}.
This allows individuals to revoke consent while ensuring that platforms remain compliant with emerging privacy regulations and societal expectations, which is the right to be forgotten~\cite{regulation2018general,song2025synthetic}. 
Such mechanisms are essential for building long-term trust between web platforms and their communities, a key requirement for achieving the broader vision of a web that serves the good of all.

\begin{figure}[t]
	\centering
	\subfigure {
		\begin{minipage}[b]{0.4\textwidth}
			\centering
			\includegraphics[width=1\textwidth]{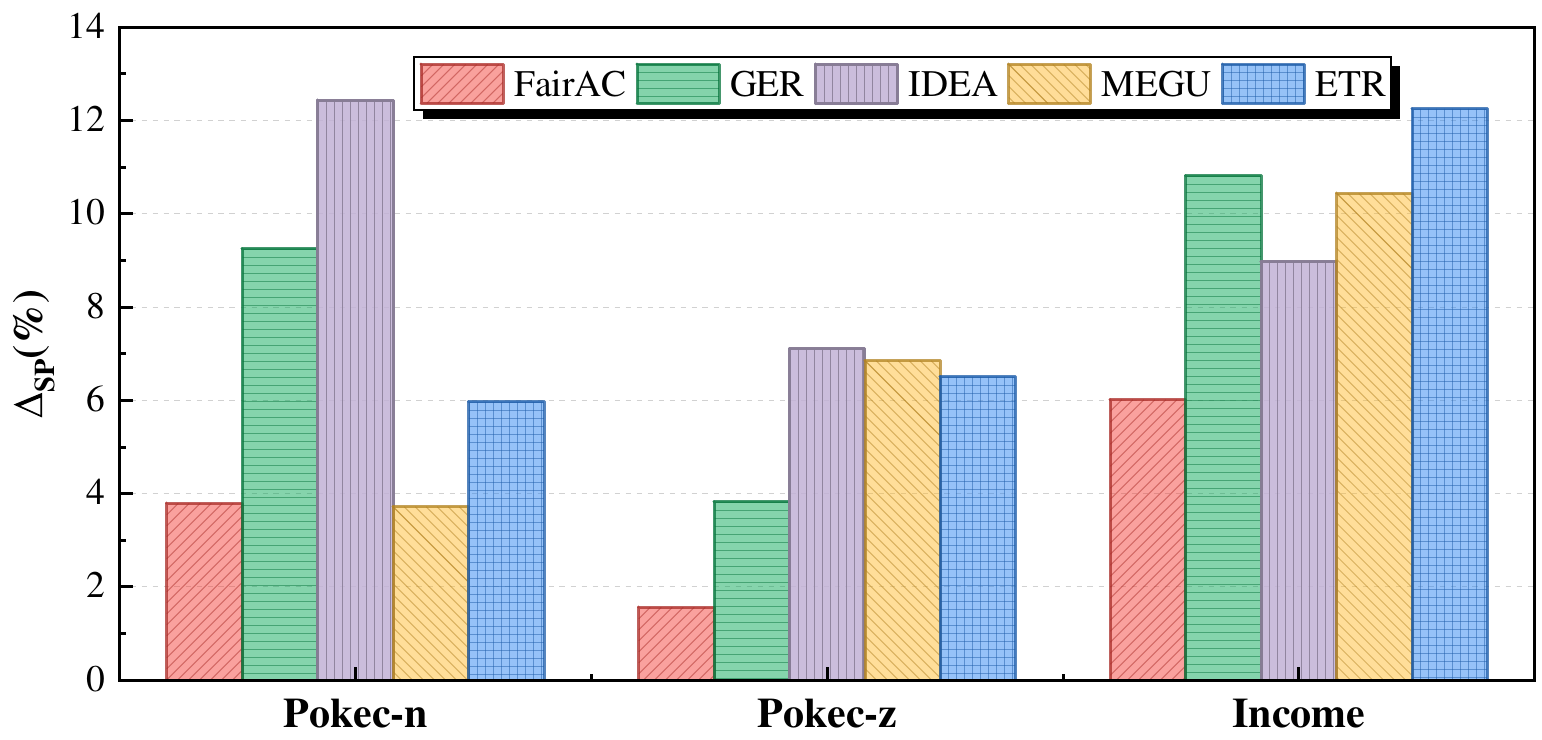}
		\end{minipage}
	}
	\subfigure {
		\begin{minipage}[b]{0.4\textwidth}
			\centering
			\includegraphics[width=1\textwidth]{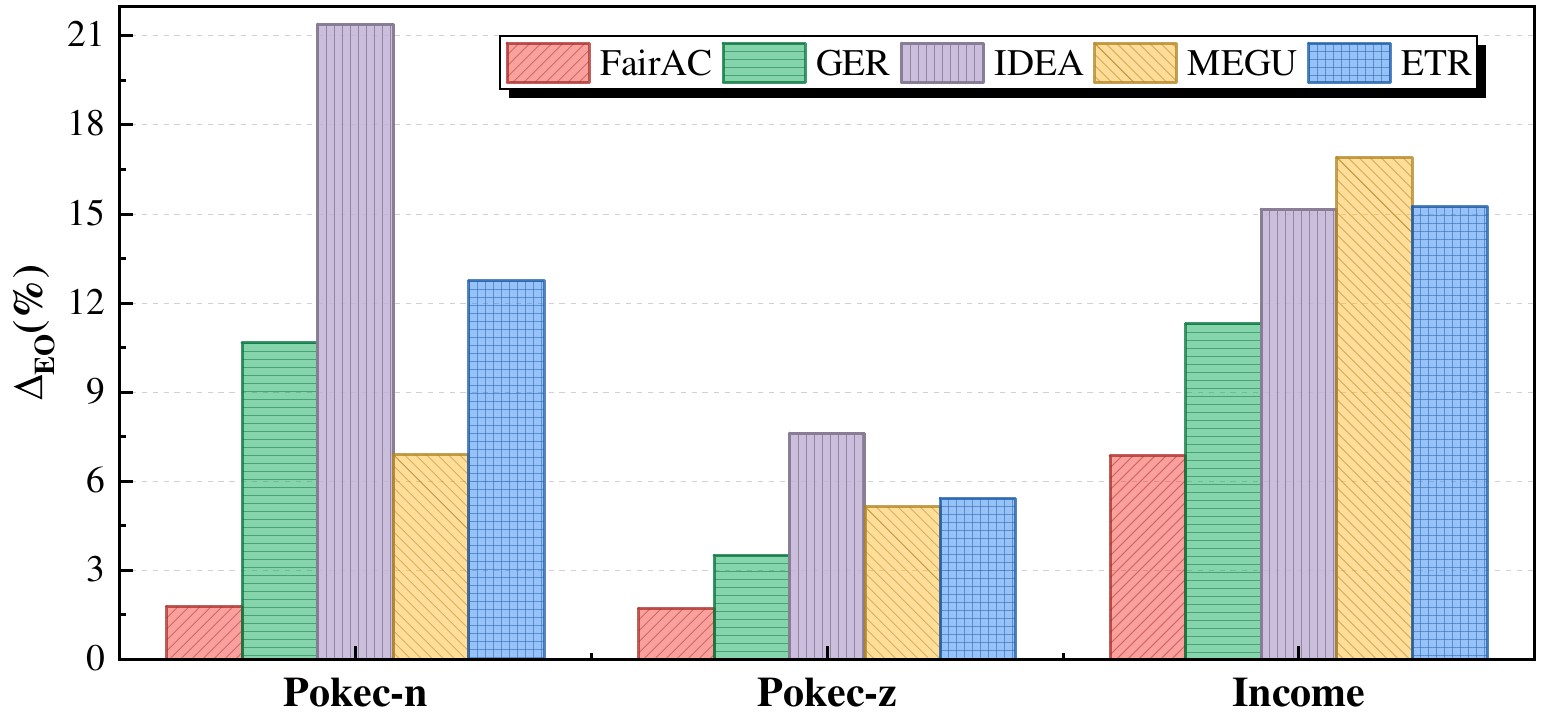}
		\end{minipage}
	}
    \caption{The $\Delta_{\text{SP}}$ and $\Delta_{\text{EO}}$ of graph unlearning methods and fairness-aware graph learning method. Lower values of $\Delta_{\text{SP}}$ and $\Delta_{\text{EO}}$ indicate better fairness performance.}
    \label{fig:background}
\end{figure}

\par Existing graph unlearning methods are primarily designed to efficiently remove explicit personal information, including sensitive attributes, from the graph. 
However, a critical question remains: \textit{do these methods preserve downstream fairness associated with those attributes?} 
In real-world web applications, algorithmic decisions must strictly adhere to principles of non-discrimination and maintain independence from protected attributes~\cite{luo2024algorithmic,XuXLCLLW22}. 
Algorithmic fairness thus provides a principled approach for evaluating whether a system treats users equitably, measured by independence criteria such as statistical parity (measured by $\Delta_\text{SP}$) and equality of opportunity (measured by $\Delta_\text{EO}$)~\cite{dong2023fairness,Oldfield0K25}.

\par To investigate this, we calculate the fairness metrics for several state-of-the-art (SOTA) graph unlearning methods (GER~\cite{chen2022graph}, IDEA~\cite{dong2024idea}, MEGU~\cite{li2024towards}, and ETR~\cite{yang2025erase}) and a fairness-aware graph learning method (FairAC~\cite{guo2023fair}) separately, as shown in Figure~\ref{fig:background}. 
We find that existing graph unlearning strategies often significantly reduce fairness, leaving models unexpectedly more sensitive to protected attributes than traditional graph learning methods. 
This critical shortcoming poses substantial ethical risks and directly threatens the inclusiveness and non-discriminatory principles that the sustainable web technologies initiative seeks to foster.

\par This challenge is further magnified when contrasting graph unlearning with fairness-aware graph learning approaches, which fundamentally rely on complete, stable graph structures and fully accessible sensitive attributes. 
Specifically, when nodes are forcibly removed during unlearning, the resulting information gaps disrupt pre-trained fairness regularizers, break critical structural cues, and severely degrade the effectiveness of built-in fairness constraints~\cite{dai2024a}. 
Such technical failures can exacerbate existing inequities as advanced web systems are increasingly deployed to serve vulnerable or marginalized communities. 
This critically undermines the broader social missions of web technologies, including applications aligned with the UN Sustainable Development Goals~\cite{xia2025graph}.

\par Motivated by these societal and technical gaps, we propose FairGU, a framework that natively integrates fairness preservation into the graph unlearning. 
FairGU handles the common scenario of incomplete sensitive attributes via a pre-trained estimator, inferring missing information privately. 
The core of our approach lies in a fairness-aware graph neural network (GNN) training paradigm that combines adversarial debiasing and covariance constraints to learn attribute-invariant node representations. 
Crucially, when fulfilling unlearning requests, FairGU employs a Fisher Information Matrix (FIM)-based importance analysis to precisely identify and dampen specialized parameters. 
This ensures the unlearning does not disrupt the learned fairness properties of the model. 
By jointly addressing sensitive attribute estimation, in-processing fairness learning, and precise unlearning, FairGU provides a robust solution for ethical graph-based web applications, promoting non-discriminatory and privacy-preserving behavior. 
The contributions are summarised as follows:

\begin{itemize}[leftmargin=0.5cm]
    \item We identify a critical limitation of existing graph unlearning methods that they insufficiently protect sensitive attributes and often degrade algorithmic fairness, revealing a previously overlooked risk in deploying unlearning techniques in socially responsible web applications.
    \item We propose a fairness-enhanced graph unlearning framework that addresses this limitation.
    This approach, combined with our fairness-control module, ensures the model's equitable behavior is maintained after unlearning, all while avoiding the privacy risks associated with reconstructing sensitive data.
    \item We conduct extensive experiments on multiple public benchmark datasets, demonstrating that our approach outperforms SOTA graph unlearning methods and fairness-aware baselines in both fairness and utility, providing a reliable and socially aligned solution for ethical and sustainable web systems.
\end{itemize}

\section{Related Work}
\subsection{Graph Unlearning}
\par Graph unlearning has become an important research direction as the need for data removal grows under privacy regulations such as the GDPR~\cite{regulation2018general}. 
It aims to remove the influence of specific data points from trained models while avoiding the high cost of full retraining~\cite{miao2025graph}.
Early work includes partition-based methods like GraphEraser (GER)~\cite{chen2022graph}, which adapts the SISA (Sharded, Isolated, Sliced, and Aggregated) paradigm~\cite{bourtoule2021machine} to graph data by introducing novel graph partition algorithms and a learning-based aggregation method, significantly reducing unlearning time while maintaining model utility~\cite{miao2025cufg}.
Subsequent efforts such as GUIDE~\cite{wang2023inductive} further optimize partitioning and aggregation strategies, but their performance remains heavily dependent on shard quality.
More recently, certified unlearning approaches have gained traction for their robustness.
IDEA~\cite{dong2024idea} stands out by offering a flexible framework for certified unlearning in GNNs, supporting four unlearning request types via an approximation method without relying on specific architectures, thus ensuring broad applicability.

\par However, many methods struggle with generalization-performan-ce trade-offs. 
For example, GNNDelete~\cite{cheng2023gnndelete} uses layer-based operators to avoid model adjustment but faces efficiency degradation with depth.
To address these challenges, MEGU~\cite{li2024towards} introduces a mutual-evolution paradigm that jointly optimizes prediction and unlearning in one unified framework. 
It achieves strong performance on feature, node, and edge unlearning tasks, while also greatly reducing both time and memory costs.
Most recently, ETR~\cite{yang2025erase} proposes a training-free method built on a theoretical framework for parameter masking. 
It follows a two-stage process, erasure and rectification, to remove unlearned samples influence, while using gradient approximation to maintain model utility, ensuring scalability and privacy without full data access.

\par Despite these advancements, existing methods predominantly prioritize utility and privacy, often overlooking algorithmic fairness.
The process of unlearning can inadvertently amplify or introduce biases against sensitive subgroups within the data, as the removal of specific nodes might disrupt the fairness properties carefully learned during the initial training.
This limitation of existing graph unlearning methods motivates the integration of fairness constraints into the unlearning process.

\subsection{Fairness-aware GNNs}
\par Parallel to developments in graph unlearning, significant research efforts have been dedicated to enhancing algorithmic fairness in GNNs, aiming to mitigate discriminatory predictions against sensitive subgroups defined by attributes such as race, gender, or region~\cite{zhang2024endowing, luo2024fairgt,XuKON25,abs-2504-21296}.
These methods primarily operate through pre-processing the data or in-processing during model training~\cite{dong2023fairness}.
Pre-processing methods like FairVGNN~\cite{wang2022improving} and Graphair~\cite{ling2023learning} aim to rectify bias at the data level by generating fair graph views or augmentations that weaken the correlation between sensitive attributes and other features.
In scenarios with missing features, FairAC~\cite{guo2023fair} jointly addresses attribute completion and unfairness, employing an attention mechanism to produce equitable node embeddings.
Conversely, in-processing methods integrate fairness constraints directly into the learning objective.
A prominent approach employs adversarial debiasing, as seen in FairGNN~\cite{dai2023learning}, which uses a sensitive attribute estimator and an adversarial loss to learn invariant node representations.  
Other methods modify the GNN architecture itself; for instance, FairSIN~\cite{yang2024fairsin} introduces a neutralization-based paradigm by incorporating fairness-facilitating features during message-passing, while FMP~\cite{jiang2024chasing} explicitly restricts the use of sensitive attributes in forward propagation.

\par A common issue in many fairness-aware GNNs is their need to generate or estimate missing sensitive attributes~\cite{luo2025fairgp}. 
This approach can help reduce bias, but it also raises privacy concerns because it reconstructs information that users may not have wanted to reveal~\cite{xiang2025use}.
More critically from the perspective of this work, these models are designed for standard training scenarios and are not inherently equipped to handle the specific challenges of data removal requests~\cite{zhang2025survey}. 
When retrofitted for unlearning (e.g., using frameworks like IDEA~\cite{dong2024idea} as a baseline adaptation), they often struggle to maintain fairness guarantees post-unlearning.
This is because their debiasing mechanisms are optimized for the original data distribution and can be disrupted by the removal of nodes, potentially reintroducing or amplifying biases.

\par This limitation underscores a significant gap in the literature: existing methods predominantly address either unlearning without considering fairness or fairness without native unlearning. 
FairGU seeks to bridge this gap by intrinsically integrating fairness preservation into the graph unlearning process.

\section{Preliminaries}
\subsection{Notations}
\par Unless stated otherwise, the following conventions are used for mathematical notions: copperplate uppercase letters (e.g., $\mathcal{A}$) represent sets, bold uppercase letters (e.g., $\mathbf{A}$) denote matrices, and bold lowercase letters (e.g., $\mathbf{a}$) signify vectors.
A graph is represented as $\mathcal{G} = (\mathcal{V}, \mathbf{A}, \mathbf{X})$, where $\mathcal{V}$ denotes the set of $n$ nodes.
$\mathbf{A} \in \{0, 1\}^{n \times n}$ is the adjacency matrix, and $\mathbf{X} \in \mathbb{R}^{n \times d}$ represents the node feature matrix, where $d$ is the dimenssion of the node feature.
Specifically, we use $|\mathcal{A}|$ to denote the number of elements in set $\mathcal{A}$.
We present group fairness for the binary label $y \in \{0,1\}$ and the sensitive attributes $s \in \{0,1\}$, where $\hat{y} \in \{0,1\}$ denotes the class label of the prediction.

\subsection{Fairness-aware Metrics}
\par \textbf{Statistical Parity}~\cite{dwork2012fairness} (i.e., Demographic Parity and Independence) requires the predictions to be independent of the sensitive attributes $s$. 
When both the predicted labels and sensitive attributes are binary, to quantify the extent of statistical parity, the $\Delta_{\text{SP}}$ is defined as follows:
\begin{equation}
  \Delta_{\text{SP}}=|\mathbb{P}(\hat{y}=1|s=0)-\mathbb{P}(\hat{y}=1|s=1)|.
\label{equ:delta_SP}
\end{equation}

\par \textbf{Equal Opportunity}~\cite{hardt2016equality} necessitates that the likelihood of an instance belonging to a positive class leading to a positive outcome should be equitable for all members within subgroups. 
For individuals with positive ground truth labels, it is necessary for positive predictions to be devoid of any dependence on sensitive attributes.
Fairness-aware GNNs prevent the allocation of unfavorable predictions to individuals who are eligible for advantageous ones solely based on their sensitive subgroup affiliation.
In particular, $\Delta_\text{EO}$ quantifies the extent of deviation in predictions from the ideal scenario where equality of opportunity is satisfied.
To quantitatively assess euqal opportunity, we employ the following metric:
\begin{equation}
    \Delta_{\text{EO}}=|\mathbb{P}(\hat{y}=1|y=1,s=0)-\mathbb{P}(\hat{y}=1|y=1,s=1)|.
\label{equ:delta_EO}
\end{equation}

\subsection{Graph Unlearning}
\par Graph unlearning extends the paradigm of machine unlearning to graph-structured data, enabling the efficient removal of specific data components (nodes or edges) from trained GNNs without the prohibitively expensive process of full retraining~\cite{chen2022graph}. 
This capability has become increasingly vital, driven by growing privacy regulations such as GDPR's "right to be forgotten"~\cite{regulation2018general}.
This is especially given the complex, interconnected nature of data in social networks and recommendation systems where graph structures naturally occur. 
The mathematical foundation of graph unlearning can be formally described as follows.

\par For the $\mathcal{G} = (\mathcal{V}, \mathbf{A}, \mathbf{X})$ with a trained GNN model $g_\omega$, the goal of graph unlearning is to remove the influence of a forget set $\mathcal{G}_f$ (a subset of $\mathcal{G}$) while maintaining performance on the retain set $\mathcal{G}_r=\mathcal{G} \setminus\mathcal{G}_f$. 
The unlearning process aims to produce a new model $g_u$ that approximates the optimal model $g_{\omega^*}$ trained solely on $\mathcal{G}_r$, which can be expressed as: 
\begin{equation}
    g_\omega\xrightarrow{\mathcal{G}_f}g_u\approx g_{\omega^*}.
\end{equation}

\par The formulation captures the essence of graph unlearning, achieving comparable results to complete retraining through efficient parameter updates rather than full model retraining. 
Requests for graph unlearning can be broadly categorized into two types: \textit{edge deletion}, which involves modifying the adjacency matrix $\mathbf{A}$ to remove specific connections, and \textit{node deletion}, which removes target nodes and their incident edges, thereby updating both $\mathcal{V}$ and $\mathbf{A}$.
The efficiency and effectiveness of this process make it a scalable solution for privacy-aware graph-based applications.

\section{The Design of FairGU}
\par
\begin{figure*}[t]
	\centering
	\includegraphics[width=0.9\textwidth]{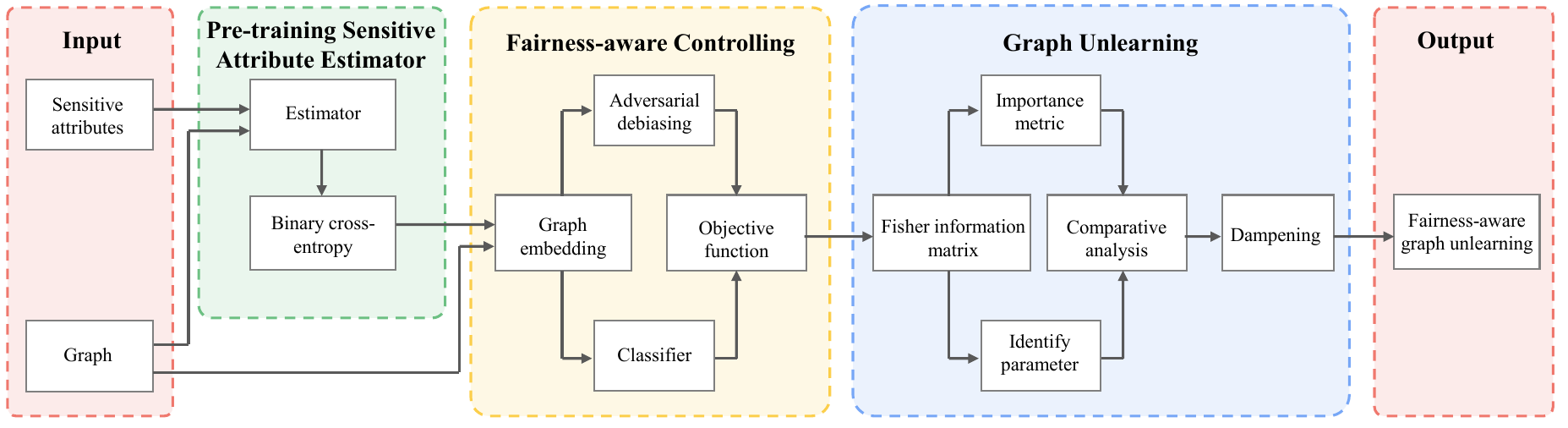}
    \caption{The illustration of FairGU.}
    \label{fig:illustration}
\end{figure*}

 In this section, we formally introduce the design of FairGU, a novel framework that seamlessly integrates fairness constraints into the graph unlearning process. 
FairGU guarantees that, following data removal operations, the model effectively eliminates the influence of specified data while strictly maintaining fairness with respect to sensitive attributes. 
The framework comprises three core components: 1) a pre-trained estimator to reconstruct missing sensitive attributes; 2) a fairness-aware graph neural network training module utilizing adversarial learning; and 3) a precise data unlearning mechanism driven by parameter importance analysis. 
These components are meticulously designed to work in synergy, enhancing both fairness and efficiency while mitigating the performance degradation often associated with privacy-preserving updates.
Figure~\ref{fig:illustration} depicts the overall model architecture, and Algorithm~\ref{alg:fairgu} details the procedural flow of the proposed framework.

\begin{algorithm}[t]
    \caption{FairGU: Fairness-aware Graph Unlearning}
    \label{alg:fairgu}
    \SetAlgoLined
    \KwIn{Graph $\mathcal{G} = (\mathcal{V}, \mathbf{A},\mathbf{X})$, sensitive attributes $s$, forget set $\mathcal{D}_f$, train set $\mathcal{D}_{train}$, hyperparameters $\alpha, \beta, \lambda, \gamma$}
    \KwOut{Unlearned model $\theta'$ with preserved fairness}
    
    \textbf{Phase 1: Fairness-aware Controlling}
    
    Initialize $f_E$ by optimizing Eq.~\eqref{equ:sae_loss} w.r.t. $\theta_E$\;
    Initialize GNN classifier $f_G$, adversary $f_A$\;
    \Repeat{convergence}{
        Obtain estimated sensitive attributes $\hat{S}$ with $f_E$\;
        Optimize the GNN classifier parameters $\theta_\mathcal{G}$, the adversary parameters $\theta_A$, and the estimator parameters $\theta_E$ by Eq.~\eqref{equ:objective function}\;
    }
    Save trained model parameters $\theta^*$\;
    
    \textbf{Phase 2: Graph Unlearning}
    
    Construct $\mathcal{G}_{\text{unlearn}}$ by removing $\mathcal{D}_f$ nodes/edges\;
    Compute FIM importance: $I_{\mathcal{D}_{\text{train}}}$, $I_{\mathcal{D}_f}$ via Eq.~\eqref{equ:importance calculation}\;
    Select parameters: $\mathcal{L}_{\text{select}} = \{i: I_{\mathcal{D}_f}(\theta_i) > \gamma \cdot I_{\mathcal{D}_{\text{train}}}(\theta_i)\}$\;
    \For{each $i \in \mathcal{L}_{\text{select}}$}{
        Compute $\text{dp}_i = \min\left(\frac{\lambda I_{\mathcal{D}_{\text{train}}}(\theta_i)}{I_{\mathcal{D}_f}(\theta_i)}, 1\right)$\;
        Update parameter: $\theta'_i = \theta_i \cdot \text{dp}_i$\;
    }
    
    \Return $\theta'$
\end{algorithm}

\subsection{Sensitive Attribute Estimator}
\label{SAE}
\par In real-world graph mining applications, sensitive attribute information is often incomplete due to inherent privacy concerns and data collection limitations. 
To address this challenge, FairGU employs a pre-training phase for sensitive attribute estimation using a graph convolutional network (GCN) based estimator $f_E$~\cite{kipf2017semi}. 
This estimator leverages both node features and graph structure to predict sensitive attributes, capturing node feature patterns while maintaining privacy preservation. 
This initial estimation step is crucial for enabling the subsequent fairness-aware training phase, which requires sensitive attribute proxies for all nodes.

\par For a given graph $\mathcal{G}=(\mathcal{V},\mathbf{A},\mathbf{X})$, the sensitive attribute estimator $f_E:\mathcal{G}\to\mathcal{S}$ is trained to predict sensitive attributes. 
The training process minimizes the binary cross-entropy loss over the set of nodes with known sensitive attributes $\mathcal{V}_S$:
\begin{equation}
    \min_{\theta_E}\mathcal{L}_E=-\frac{1}{|\mathcal{V}_S|}\sum{v\in\mathcal{V}_S}[s_v\log\hat{s}_v+(1-s_v)\log{(1-\hat{s}_v)}],
    \label{equ:sae_loss}
\end{equation}
where $\hat{s}_v$ is the predicted sensitive attribute of node $v\in\mathcal{V}_S$ by $f_E$ and $\theta_E$ is the set of parameters of $f_E$. 
This pre-training phase ensures that the estimator learns to approximate the original sensitive attribute distribution without generating synthetic data, aligning with privacy preservation principles while enabling fairness-aware learning in scenarios with incomplete sensitive attributes. 
The resulting estimated attributes $\hat{S}$ are then used as proxies for the sensitive attributes of all nodes in the subsequent fairness-aware GNN training.

\subsection{Fairness-aware Controlling}
\par FairGU utilizes a fairness-aware GNN architecture that integrates graph structural information and node features while enforcing fairness constraints. 
The primary goal of this control mechanism is to generate highly effective node embeddings that are statistically independent of the sensitive attribute.
The framework consists of three main components:
\begin{itemize} [leftmargin=0.5cm]
    \item A GNN classifier $f_\mathcal{G}$ for node classification.
    \item A sensitive attribute estimator $f_{E}$ (pre-trained as in Section \ref{SAE}).
    \item An adversary $f_{A}$ for adversarial debiasing.
\end{itemize}

\par The GNN classifier $f_\mathcal{G}$ takes the graph $\mathcal{G}$ as input and produces node representations. 
For a node $v$, after $K$ layers of aggregation, the representation $h_v$ is obtained as:
\begin{equation}
h_v=f_\mathcal{G}^{(K)}\left(x_v,\mathcal{N}_v^{(K)}\right), 
\end{equation}
where $\mathcal{N}_v^{(K)}$ denotes the $K$-hop neighborhoods of $v$. 
The prediction $\hat{y}_v$ is then computed through a linear classification layer with sigmoid activation:
\begin{equation}
\hat{y}_v=\sigma\left(h_v\cdot w\right),
\end{equation}
where $w\in\mathbb{R}^d$ is the weight vector. 
The classification loss $\mathcal{L}_C$ is the binary cross-entropy loss over the labeled node set $\mathcal{V}_L$:
\begin{equation}
\min_{\theta_\mathcal{G}}\mathcal{L}_C=-\frac{1}{|\mathcal{V}_L|}\sum_{v\in\mathcal{V}_L}\left[y_v\log\hat{y}_v+(1-y_v)\log(1-\hat{y}_v)\right],
\end{equation}
where $\theta_\mathcal{G}$ represents the parameters of $f_\mathcal{G}$.

\par To ensure fairness, FairGU incorporates adversarial debiasing. 
The adversary $f_{A}$, which utilizes a linear classifier, aims to predict the sensitive attribute from the node representation $h_{v}$, while $f_\mathcal{G}$ aims to learn representations that deceive $f_{A}$. 
This min-max game is formulated as:
\begin{equation}
    \begin{aligned}
    \min_{\theta_{\mathcal{G}}}\max_{\theta_{A}}\mathcal{L}_{A}&=\mathbb{E}_{{h}\sim p(h|\hat{s}=1)}[\log(f_A({h}))]\\&+\mathbb{E}_{h\sim p(h|\hat{s}=0)}[\log(1-f_A(h))],
    \end{aligned}    
\end{equation}
where $\hat{s}\in\hat{\mathcal{S}}$ is the estimated sensitive attribute (from $f_{E}$) for nodes in $\mathcal{V}$, and $\theta_{A}$ is the parameters of $f_{A}$. 
This adversarial training encourages the representations $h_{v}$ to be independent of the sensitive attributes.

\par Additionally, a covariance constraint $\mathcal{L}_{R}$ is applied to directly minimize the dependence between predictions and sensitive attributes:
\begin{equation}
    \mathcal{L}_{R}=|\mathrm{Cov}(\hat{s},\hat{y})|=|\mathbb{E}[(\hat{s}-\mathbb{E}[\hat{s}])(\hat{y}-\mathbb{E}[\hat{y}])]|,
\end{equation}
where $\hat{y}$ is the predicted label and $\hat{s}$ is the estimated sensitive attribute. 
This constraint ensures statistical parity by minimizing the covariance between predictions and sensitive attributes.

\par The overall objective function combines the classification loss, estimator loss, adversarial loss, and covariance constraint:
\begin{equation}
    \min_{\theta_{\mathcal{G}},\theta_E}\max_{\theta_A}\mathcal{L}_C+\mathcal{L}_E+\alpha\mathcal{L}_R-\beta\mathcal{L}_A,
    \label{equ:objective function}
\end{equation}
where $\alpha$ and $\beta$ are hyperparameters controlling the contribution of the covariance constraint and adversarial debiasing, respectively. 
The training follows an alternating optimization strategy: first, update the generator components ($f_{\mathcal{G}}$ and $f_E$) to minimize $\mathcal{L}_C+\mathcal{L}_E+\alpha\mathcal{L}_R-\beta\mathcal{L}_A$ with $f_A$ fixed; then update the adversary $f_A$ to maximize $\mathcal{L}_A$ with the generator components fixed. 
This process ensures that the model learns fair node representations without sacrificing predictive performance.

\subsection{Graph Unlearning}
\par FairGU employs FIM-based parameter importance analysis to perform precise data unlearning. 
The Fisher Information Matrix (FIM) serves as a fundamental tool for quantifying the sensitivity of model parameters with respect to input samples~\cite{guo2019certified}.
It captures the amount of information that an observable random variable carries about unknown parameters, providing a principled measure of parameter importance.
Formally, for a distribution $p(y|x,w)$, the FIM is defined as the expected value of the outer product of the score function~\cite{kay1993fundamentals}:
\begin{equation}
    \begin{aligned}
    F_{D}=&\mathbb{E}_{x,y}\left[-\nabla_{w}^{2}\log p\left(y|x,w\right)\right]\\
    =&\mathbb{E}_{x,y}\left[\nabla_{w}\log p\left(y|x,w\right)\nabla_{w}\log p\left(y|x,w\right)^{T}\right].
    \end{aligned}
\end{equation}

\par Notably, under certain regularity conditions, the FIM is equivalent to the negative expected Hessian of the log-likelihood function~\cite{pawitan2001all}, establishing a connection between first-order and second-order derivatives of the loss function.
However, computing the full FIM is computationally prohibitive for large-scale models due to its quadratic size in the number of parameters.
To address this challenge, we follow established practices in the literature~\cite{kirkpatrick2017overcoming} and utilize the diagonal of the FIM, which can be efficiently computed using first-order derivatives while still providing a reliable approximation of parameter importance.
The $i$-th diagonal element $F_{\mathcal{D},ii}$ represents the importance of the $i$-th parameter with respect to dataset $\mathcal{D}$, offering a computationally tractable alternative to the full matrix.

\par For graph data, the importance calculation incorporates structural information, and the importance metric for a dataset $\mathcal{D}$ (which can be the entire training set $\mathcal{D}_{train}$, the forget set $\mathcal{D}_{f}$, or the retain set  $\mathcal{D}_{r}$) on parameters $\theta$ is computed as: 

\begin{equation}
    I_{\mathcal{D}}(\theta)=\frac{1}{|\mathcal{D}|}\sum_{v\in\mathcal{D}}\left(\frac{\partial\mathcal{L}(f_\theta(\mathbf{X},\mathbf{A})[v],y_v)}{\partial\theta}\right)^2,
    \label{equ:importance calculation}
\end{equation}
where $\mathcal{L}$ is the loss function, and $f_\theta(\mathbf{X},\mathbf{A})$ represents the GNN model taking $\mathbf{X}$ and $\mathbf{A}$ as input. 
The importance for the entire training set $I_{\mathcal{D}_{train}}(\theta)$, is computed once after the initial training and stored. 
This allows the original training data to be discarded, enhancing privacy and reducing storage requirements. 
For each unlearning request, the importance for the forget set $I_{\mathcal{D}_f}(\theta)$, is calculated. 
The core of our method lies in the comparative analysis of FIM-based importance metrics. 
We identify parameters that exhibit significantly higher importance for the forget set $\mathcal{D}_{f}$ compared to the overall training set $\mathcal{D}_{train}$:
\begin{equation}
    \mathcal{L}_{select}=\{i:I_{\mathcal{D}_f} (\theta_i)>\gamma I_{\mathcal{D}_{train}}(\theta_i)\},
    \label{equ:loss select}
\end{equation}
where the hyperparameter $\gamma$ controls the selectivity of the process: higher values protect more parameters, preserving generalized knowledge, while lower values enable more aggressive unlearning of specific patterns.

\par Instead of pruning selected parameters (which could harm overall model performance), for selected parameters, we apply a targeted dampening factor proportional to their relative specialization towards the forget set. 
The dampening factor $\text{dp}_{i}$ is computed as: 
\begin{equation}
    \text{dp}_i=\min\left(\frac{\lambda I_{\mathcal{D}_{train}}(\theta_i)}{I_{\mathcal{D}_f}(\theta_i)},1\right),
    \label{equ:dampening factor}
\end{equation}
where the parameter is then updated using a multiplicative rule: 
\begin{equation}
\theta_i^{\prime}=\theta_i\text{dp}_i.
\end{equation}
\par The hyperparameter $\lambda$ controls the strength of the dampening. 
This formulation ensures that parameters deemed highly specialized for the forget set are dampened more severely ($\text{dp}_i\to0$), while those of general utility are preserved ($\text{dp}_{i}\to1$). 
The $\min(\cdot,1)$ operation ensures the update only reduces the parameter's magnitude, preventing unintended amplification and maintaining training stability. 
Parameters not selected for modification remain unchanged. 

\par Given structural dependencies in graph data, FairGU constructs an updated graph structure $\mathcal{G}_{unlearn}=(\mathcal{V^{\prime}},\mathbf{A}^{\prime}, \mathbf{X}^{\prime})$, where:
\begin{itemize}
\item $\mathcal{V}^{\prime}=\mathcal{V}\setminus\mathcal{V}_{forget}$: removes nodes to be forgotten.
\item $\mathbf{A}^{\prime}$: updates the adjacency matrix by removing edges connected to forgotten nodes.
\item $\mathbf{X}^{\prime}$: updates the node feature matrix accordingly 
\end{itemize}
\par The FIM-based importance calculation and subsequent dampening are performed using this updated graph $\mathcal{G}_{unlearn}$, ensuring semantic correctness in graph data unlearning tasks.

\section{Experiments}

\subsection{Datasets}
\par In this study, we evaluate FairGU on three real-world datasets: Income, Pokec-z, and Pokec-n. 
These datasets are chosen for their diverse domains, inherent sensitive attributes, and relevance to practical scenarios where data removal requests may arise due to privacy concerns or regulatory compliance. 
The key statistics of the datasets are summarized in Table~\ref{tab:dataset}. 

\begin{table}[t]
    \centering
    \caption{Statistics of the three real-world datasets. }
    \renewcommand{\arraystretch}{0.8}
    \begin{tabular}{ccccc}\toprule
         Datasets&  Nodes&  Edges&  Sensitive attributes& Label\\\midrule
         Pokec-z&  $67,797$&   $882,765$&  Region/Gender&  Field\\
         Pokec-n&  $66,569$&   $729,129$&  Region/Gender&  Field\\
         Income&  $14,821$&  $100,483$&  Race&  Income\\ \bottomrule
    \end{tabular}
    \label{tab:dataset}
\end{table}

\begin{table*}[t]
    \centering
    \caption{Performance comparison between FairGU and baseline methods on Pokec-n, Pokec-z, and Income datasets. 
    ACC represents accuracy, $\Delta_{\text{SP}}$ represents statistical parity difference, and $\Delta_{\text{EO}}$ represents equal opportunity difference. 
    Lower values of $\Delta_{\text{SP}}$ and $\Delta_{\text{EO}}$ indicate better fairness performance. 
    The best results are highlighted in \textcolor{red}{\textbf{red and bold}}.}
    \label{tab:results_comparison}
    \renewcommand{\arraystretch}{0.8}
    \begin{tabular}{lccccccccc}
    \toprule
    \multirow{2}{*}{Method} & \multicolumn{3}{c}{Pokec-n} & \multicolumn{3}{c}{Pokec-z} & \multicolumn{3}{c}{Income} \\
    \cmidrule(lr){2-4} \cmidrule(lr){5-7} \cmidrule(lr){8-10}
     & ACC (\%) $\uparrow$ & $\Delta_{\text{SP}}$ (\%) $\downarrow$ & $\Delta_{\text{EO}}$ (\%) $\downarrow$ & ACC (\%) $\uparrow$ & $\Delta_{\text{SP}}$ (\%) $\downarrow$ & $\Delta_{\text{EO}}$ (\%) $\downarrow$ & ACC (\%) $\uparrow$ & $\Delta_{\text{SP}}$ (\%) $\downarrow$ & $\Delta_{\text{EO}}$ (\%) $\downarrow$ \\
    \midrule
    GER (CCS'22) & $66.81_{\pm 1.07}$ & $9.25_{\pm 2.34}$ & $10.67_{\pm 2.86}$ & $66.43_{\pm 0.63}$ & $3.83_{\pm 1.09}$ & $3.51_{\pm 1.64}$ & $79.34_{\pm 0.11}$ & $10.82_{\pm 1.11}$ & $11.30_{\pm 1.49}$ \\
    IDEA (KDD'24) & $65.27_{\pm 0.28}$ & $12.44_{\pm 0.59}$ & $21.39_{\pm 2.79}$ & $64.02_{\pm 0.54}$ & $7.11_{\pm 1.95}$ & $7.62_{\pm 1.34}$ & $79.48_{\pm 0.03}$ & $8.97_{\pm 0.19}$ & $15.16_{\pm 0.66}$ \\
    MEGU (AAAI'24) & $67.03_{\pm 1.14}$ & $3.72_{\pm 0.89}$ & $6.90_{\pm 1.02}$ & $65.79_{\pm 0.54}$ & $6.85_{\pm 1.89}$ & $5.13_{\pm 1.77}$ & $78.84_{\pm 0.06}$ & $10.44_{\pm 0.71}$ & $16.91_{\pm 2.03}$ \\
    ETR (AAAI'25) & $62.84_{\pm 0.48}$ & $5.97_{\pm 0.66}$ & $12.75_{\pm 1.18}$ & $65.87_{\pm 0.89}$ & $6.51_{\pm 1.48}$ & $5.43_{\pm 1.18}$ & \textcolor{red}{$\mathbf{81.59_{\pm 0.09}}$} & $12.25_{\pm 0.28}$ & $15.25_{\pm 0.86}$ \\
    \midrule
    FairGNN (TKDE'23) & $65.27_{\pm 0.23}$ & $5.28_{\pm 1.47}$ & $7.72_{\pm 2.05}$ & $64.58_{\pm 0.25}$ & $5.95_{\pm 1.38}$ & $5.12_{\pm 1.55}$ & $79.19_{\pm 0.24}$ & $7.27_{\pm 0.78}$ & $28.30_{\pm 2.17}$ \\
    FairAC (ICLR'23) & $62.21_{\pm 1.47}$ & $3.79_{\pm 0.43}$ & $1.78_{\pm 0.77}$ & $63.31_{\pm 2.84}$ & $1.57_{\pm 0.51}$ & $1.72_{\pm 0.45}$ & $76.58_{\pm 1.41}$ & $6.01_{\pm 0.85}$ & $6.85_{\pm 1.27}$ \\
    FairSIN (AAAI'24) & $66.32_{\pm 0.45}$ & $4.86_{\pm 0.65}$ & $3.89_{\pm 2.41}$ & $63.49_{\pm 2.57}$ & $2.52_{\pm 1.78}$ & $4.43_{\pm 1.34}$ & $80.54_{\pm 0.63}$ & $9.86_{\pm 1.46}$ & $12.00_{\pm 1.12}$ \\
    \midrule
    \textbf{FairGU} & \textcolor{red}{$\mathbf{67.05_{\pm 0.77}}$} & \textcolor{red}{$\mathbf{0.74_{\pm 0.39}}$} & \textcolor{red}{$\mathbf{1.54_{\pm 0.78}}$} & \textcolor{red}{$\mathbf{67.63_{\pm 0.77}}$} & \textcolor{red}{$\mathbf{1.12_{\pm 1.07}}$} & \textcolor{red}{$\mathbf{0.58_{\pm 0.30}}$} & $80.40_{\pm 0.03}$ & \textcolor{red}{$\mathbf{0.42_{\pm 0.07}}$} & \textcolor{red}{$\mathbf{1.33_{\pm 0.39}}$} \\
    \bottomrule
    \end{tabular}
\end{table*}

\begin{itemize}[leftmargin=0.5cm]
    \item \textbf{Income}~\cite{asuncion2007uci}: The Income dataset is derived from the Adult Data Set.
    It represents a social network where each node corresponds to an individual, and edges are established based on similarity criteria such as demographic or socioeconomic features.
    The sensitive attribute is race, and the primary classification task is to predict whether an individual's annual income exceeds \$$50,000$.
    This dataset is highly relevant for fairness-aware unlearning studies, as income prediction models can potentially perpetuate or amplify biases related to racial subgroups.
    The need to forget specific user data may stem from privacy laws or individual requests to remove financial information. 
    \item \textbf{Pokec-z} and \textbf{Pokec-n}~\cite{takac2012data}: Pokec-z and Pokec-n are constructed by sampling users from two different Slovakian provinces. 
    The primary task for both datasets is to predict a user's working field. 
    These datasets include two sensitive attributes: region (the province of residence) and gender. 
    Social networks like Pokec are typical scenarios where users might exercise their "right to be forgotten" under regulations like GDPR, requesting the removal of their profiles, associations, and related data from trained models. 
    The large scale of these graphs also tests the scalability of unlearning methods.
\end{itemize}

\par To align with standard evaluation practices in graph unlearning literature~\cite{chen2022graph,dong2024idea}, we preprocess the datasets as follows. 
For datasets with multi-class labels, we binarize the prediction task by mapping labels greater than $1$ to a single class.
For each dataset, we randomly split the graph into two disjoint parts, with $80$\% of nodes used for training the GNN model and the remaining $20$\% reserved for evaluating model utility. 
For the unlearning process, a portion of nodes within the training set is designated as the forget set ($\mathcal{D}_{f}$),  while the remainder constitutes the retain set ($\mathcal{D}_{r}$).
Specifically, for our experiments, we perform a $5$\% node deletion task on each dataset, along with the removal of associated edges.

\subsection{Baselines}
\par In our experiments, we compare FairGU against a comprehensive set of SOTA baselines, which are categorized into two groups: 1) graph unlearning methods that focus on efficiently removing data influences from trained models, and 2) fairness-aware GNNs that aim to mitigate bias in graph learning. 
For the fairness-aware GNNs, since they are not originally designed for graph unlearning, we adapt each model using the IDEA framework~\cite{dong2024idea}, to enable a fair comparison under unlearning scenarios, ensuring consistency in handling removal requests.

\par We list four graph unlearning baselines:
\begin{itemize} [leftmargin=0.5cm]
    \item \textbf{GER (GraphEraser)}~\cite{chen2022graph} introduces novel graph partition algorithms and a learning-based aggregation method to address unlearning requests efficiently, achieving significant reductions in unlearning time and improvements in model utility.
    \item \textbf{IDEA}~\cite{dong2024idea} is a flexible framework for certified unlearning in GNNs, providing theoretical guarantees for information removal across various unlearning types and supporting diverse GNN architectures without requiring specific designs.
    \item \textbf{MEGU}~\cite{li2024towards} proposes a mutual evolution paradigm that simultaneously optimizes predictive and unlearning capabilities through a unified training framework, demonstrating strong performance on feature, node, and edge unlearning tasks with substantial efficiency gains.
    \item \textbf{ETR}~\cite{yang2025erase} is a training-free unlearning approach that operates in two stages—erasure and rectification—by editing model parameters to eliminate unlearned sample influences and approximating gradients to preserve utility, offering scalability and privacy benefits.
\end{itemize}

\par We use three fairness-aware GNN baselines with IDEA adaptation:
\begin{itemize} [leftmargin=0.5cm]
    \item \textbf{FairGNN}~\cite{dai2023learning} addresses bias in GNNs by leveraging graph structures and limited sensitive attribute information. 
    It uses adversarial training and data generation to debias node embeddings while maintaining accuracy.
    \item \textbf{FairAC}~\cite{guo2023fair} jointly tackles attribute completion and unfairness in graphs with missing attributes. 
    It employs an attention mechanism to complete missing features and mitigate feature and topological unfairness, producing fair embeddings for downstream tasks.
    \item \textbf{FairSIN}~\cite{yang2024fairsin} introduces a neutralization-based paradigm for fairness, incorporating fairness-facilitating features into node representations to statistically neutralize sensitive bias. 
    It emphasizes heterogeneous neighbors' features to achieve fairness without filtering out non-sensitive information.
\end{itemize}

\par These baselines represent the current SOTA in graph unlearning and fairness-aware graph learning. 
By adapting the fairness-aware GNNs with IDEA's unlearning capability, we ensure a comprehensive and fair evaluation of FairGU against methods that address both fairness and unlearning challenges.

\begin{table}[t]
    \centering
    \caption{Comparison of MIA AUC Scores (\%) for baseline methods and FairGU. The best results are \textcolor{red}{red and bold-faced}. A value closer to $50$\% is preferred, as it corresponds to the behavior of random guessing. The runner-ups are \textcolor{blue}{\underline{blue and underlined}}.}
    \label{tab:mia_auc_comparison}
    \tabcolsep=0.3cm
    \renewcommand{\arraystretch}{0.8}
    \begin{tabular}{lcccc}
    \toprule
    \textbf{Method} & \textbf{Pokec-n} & \textbf{Pokec-z} & \textbf{Income} \\
    \midrule
    GER     & $53.1 \pm 1.1$ & $50.8 \pm 0.5$ & \textcolor{blue}{\underline{$50.7\pm 0.4$}} \\
    IDEA    & $52.3 \pm 1.6$ & $51.7 \pm 0.5$ & $52.9 \pm 1.6$ \\
    MEGU    & $52.1 \pm 1.4$ & \textcolor{red}{$\mathbf{50.6 \pm 0.4}$} & $50.9 \pm 2.8$ \\
    ETR     & \textcolor{red}{$\mathbf{51.0 \pm 0.8}$} & $50.9 \pm 0.7$ & $51.1 \pm 0.5$ \\
    \midrule
    \textbf{FairGU} & \textcolor{blue}{\underline{$51.2 \pm 0.9$}} & \textcolor{blue}{\underline{$50.7 \pm 1.0$}} & \textcolor{red}{$\mathbf{50.2 \pm 0.3}$} \\
    \bottomrule
    \end{tabular}
\end{table}

\subsection{Experimental Setup}
\par All experiments are conducted on a workstation equipped with an NVIDIA GeForce RTX $4070$ GPU. 
The software environment includes Python $3.8.10$, PyTorch $1.13.0$, and CUDA $11.7.0$.
For the sensitive attribute estimator $f_E$, we deploy a one-hidden-layer Graph Convolutional Network (GCN) with a hidden dimension of 128. 
The adversary $f_A$ utilizes a linear classifier.
To validate the general applicability of our framework across different GNN architectures, we employ both GCN and Graph Attention Network (GAT) as backbones for the fairness-aware GNN classifier.
Specifically, we use a two-layer GCN with a hidden dimension of $128$ and a two-layer GAT with a hidden dimension of $64$, respectively.
Hyperparameters are selected based on performance on the validation set. The fairness constraint weights $\alpha$ and $\beta$ in Equation~\eqref{equ:objective function} are tuned within the ranges of [$0.0001$, $1$] and [$1$, $100$], respectively.
For the unlearning phase, the selectivity threshold $\gamma$ in Equation~\eqref{equ:loss select} is optimized within [$0.1$, $100$], and the dampening strength $\lambda$ in Equation~\eqref{equ:dampening factor} is selected from [$0.1$, $5$].
The specific values for these hyperparameters are determined separately for each dataset based on validation performance.
To ensure statistical reliability, each experiment related to model utility is repeated $10$ times, and we report the mean and standard deviation of the results.

\subsection{Comparison Results}
\par In this section, we present a comprehensive evaluation of FairGU against SOTA baselines under a $5$\% node unlearning scenario across three real-world datasets: Pokec-n, Pokec-z, and Income. 
The comparison focuses on four key aspects: predictive accuracy (ACC), fairness metrics (statistical parity difference $\Delta_{\text{SP}}$ and equal opportunity difference $\Delta_{\text{EO}}$), and privacy preservation measured by Membership Inference Attack (MIA) AUC scores. 
The results shown in Table~\ref{tab:results_comparison} demonstrate that FairGU consistently outperforms existing methods in balancing utility and fairness.
Specifically, it effectively reduces algorithmic bias while retaining competitive predictive performance and privacy protection.

\par For predictive accuracy, FairGU maintains competitive node classification accuracy across all datasets, achieving performance on par with or superior to the best baselines. 
For instance, on Pokec-n and Pokec-z, FairGU delivers high accuracy that rivals leading graph unlearning methods like MEGU, while on Income, it closely matches the accuracy of fairness-aware approaches such as FairSIN. 
Although on the Income dataset, FairGU's accuracy is slightly lower than that of ETR, this minor sacrifice in accuracy is offset by a substantial enhancement in fairness. 
This trade-off highlights FairGU's ability to strike an optimal balance between utility and fairness, ensuring that the model not only retains high predictive performance but also prioritizes ethical considerations by minimizing biases.
This indicates that FairGU effectively preserves model utility even after unlearning, without significant degradation in prediction quality.

\par For fairness performance, fairGU significantly enhances fairness by reducing both statistical parity and equal opportunity disparities. 
Compared to all baselines, including fairness-aware GNNs adapted for unlearning, FairGU achieves the lowest $\Delta_{\text{SP}}$ and $\Delta_{\text{EO}}$ values, indicating a substantial reduction in bias. 
For example, on Pokec-n, FairGU's fairness metrics are markedly lower than those of MEGU and FairGNN, demonstrating improvements of over $75$\% in statistical parity and $77$\% in equal opportunity. 
Similarly, on Pokec-z and Income, FairGU outperforms methods like FairAC and FairSIN by wide margins, minimizing discrimination against sensitive subgroups. 
This underscores FairGU's capability to promote fairness without compromising on accuracy.

\par To evaluate the effectiveness of graph unlearning, we employ the SOTA threat model MIA-Graph~\cite{olatunji2021membership} to conduct membership inference attacks (MIA), which aim to infer whether a node was part of the model's training set. 
The effectiveness of unlearning is measured by the AUC score of the MIA.
A lower AUC score indicates better unlearning performance, with an AUC of $50$\% representing the ideal case where the attack model performs no better than random guessing, thus signifying that the unlearned data's membership information has been effectively erased.
The results in Table ~\ref{tab:mia_auc_comparison} show that FairGU provides robust privacy protection, with attack accuracy close to 50\%, which corresponds to random guessing. 
This indicates that adversaries cannot reliably infer membership, highlighting FairGU's effectiveness in safeguarding sensitive information. 
In contrast, some baselines exhibit higher MIA AUC scores, suggesting greater vulnerability to MIA. 
FairGU's privacy performance is comparable to retraining from scratch and superior to most graph unlearning baselines, aligning with its design goal of avoiding synthetic data generation.

\par Overall, FairGU emerges as a holistic solution that excels in accuracy, fairness, and privacy. It effectively addresses the challenges of node unlearning with missing sensitive attributes, outperforming specialized baselines in each category. 
The consistent gains across datasets and metrics validate FairGU's robustness and scalability, making it suitable for real-world applications where ethical and privacy concerns are paramount. 
By integrating fairness-aware training with FIM-based parameter analysis, FairGU achieves SOTA results while mitigating secondary privacy risks and ensuring durable fairness guarantees post-unlearning.

\subsection{Ablation Study}
\par To comprehensively evaluate the contributions of key components in FairGU, we conduct an ablation study focusing on two critical aspects: the pre-training mechanism for the sensitive attribute estimator and the fairness-aware conttroling architecture. 
The study compares the full FairGU model against two variants: 
(1) FairGU without pre-training the sensitive attribute estimator (denoted as FairGU w/o SAE), where the estimator is trained concurrently during the main training phase rather than being pre-trained, leading to reduced accuracy in sensitive attribute prediction; and 
(2) FairGU without the fairness-aware controlling, replaced by a standard GCN backbone (denoted as FairGU w/o FC), to assess the impact of the fairness-specific design. 
Experiments are conducted across multiple datasets including Income, Pokec-z, and Pokec-n under a $5$\% node unlearning scenario, with results summarized in Table~\ref{tab:ablation_results}.

\begin{table}[t]
    \centering
    \caption{Ablation study results of FairGU variants across datasets. }
    \label{tab:ablation_results}
    \renewcommand{\arraystretch}{0.8}
    \begin{tabular}{lrrrr}
        \toprule
        Dataset & Metric & FairGU & w/o SAE & w/o FC \\
        \midrule
        Pokec-n & ACC (\%) $\uparrow$ & ${67.05_{\pm 0.77}}$ & $66.21_{\pm 0.84}$ & \textcolor{red}{$\mathbf{67.78_{\pm 0.49}}$} \\
        & $\Delta_{\text{SP}}$ (\%) $\downarrow$ & \textcolor{red}{$\mathbf{0.74_{\pm 0.39}}$} & $3.86_{\pm 1.61}$ & $5.17_{\pm 1.56}$ \\
        & $\Delta_{\text{EO}}$ (\%) $\downarrow$ & \textcolor{red}{$\mathbf{1.54_{\pm 0.78}}$} & $4.20_{\pm 0.74}$ & $8.69_{\pm 1.08}$ \\
        \midrule
        Pokec-z & ACC (\%) $\uparrow$ & \textcolor{red}{$\mathbf{67.63_{\pm 0.77}}$} & $66.57_{\pm 0.85}$ & $66.88_{\pm 0.79}$ \\
        & $\Delta_{\text{SP}}$ (\%) $\downarrow$ & \textcolor{red}{$\mathbf{1.12_{\pm 1.07}}$} & $2.14_{\pm 1.49}$ & $7.05_{\pm 1.75}$ \\
        & $\Delta_{\text{EO}}$ (\%) $\downarrow$ & \textcolor{red}{$\mathbf{0.58_{\pm 0.30}}$} & $3.42_{\pm 0.97}$ & $5.42_{\pm 1.47}$ \\
        \midrule
        Income & ACC (\%) $\uparrow$ & $80.40_{\pm 0.03}$ & $79.83_{\pm 0.16}$ & \textcolor{red}{$\mathbf{80.88_{\pm 0.11}}$} \\
        & $\Delta_{\text{SP}}$ (\%) $\downarrow$ & \textcolor{red}{$\mathbf{0.42_{\pm 0.07}}$} & $4.17_{\pm 2.33}$ & $7.27_{\pm 0.78}$ \\
        & $\Delta_{\text{EO}}$ (\%) $\downarrow$ & \textcolor{red}{$\mathbf{1.33_{\pm 0.39}}$} & $9.52_{\pm 1.62}$ & $27.28_{\pm 1.45}$ \\
        \bottomrule
    \end{tabular}
\end{table}

\par The ablation study results demonstrate that both components are essential for achieving an optimal balance between fairness and utility.
In particular, the FairGU w/o FC variant operates without the regularization penalties designed to enforce fairness, allowing it to focus exclusively on minimizing classification error.
As a result, this variant occasionally achieves slightly higher predictive accuracy on certain datasets (e.g., Pokec-n and Income).
However, this marginal gain in utility is obtained at a severe ethical cost.
Without the explicit fairness constraints, the unlearning process fails to mitigate discriminatory patterns, leading to $\Delta_{\text{SP}}$ and $\Delta_{\text{EO}}$ values that are substantially higher than those of the full FairGU model.
This finding underscores that sacrificing fairness for negligible improvements in accuracy is not desirable in practice, especially in socially sensitive web applications.
Similarly, for the FairGU w/o SAE variant, the absence of pre-training results in less accurate estimations of sensitive attributes, which exacerbates fairness disparities.
This is reflected in increased $\Delta_{\text{SP}}$ and $\Delta_{\text{EO}}$ across all datasets, alongside a modest decrease in accuracy.
The consistency of these trends confirms the robustness of FairGU's design.
Ultimately, the full FairGU model consistently maintains the best fairness performance ($\Delta_{\text{SP}}$ and $\Delta_{\text{EO}}$) while preserving competitive accuracy, highlighting that integrating these fairness-aware components is crucial for achieving bias mitigation without severely compromising utility.

\par In summary, the ablation study underscores that pre-training the sensitive attribute estimator and incorporating a fairness-aware GNN are indispensable for FairGU's fairness performance. 
The concurrent training of the estimator without pre-training compromises fairness, while replacing the fairness-aware GNN, though occasionally boosting accuracy, leads to unacceptable fairness degradation. 
These findings reinforce the necessity of both components in practical applications where fairness is paramount.

\section{Conclusion}
\par In this paper, we propose FairGU, a new fairness-aware graph unlearning framework. 
It tackles the key challenge of applying fairness constraints during the data removal process in graph neural networks, especially when sensitive attributes are missing.
FairGU ensures that after unlearning, the model removes the influence of the specified data while still maintaining fairness with respect to sensitive attributes. 
It achieves this through three components: a pre-trained sensitive-attribute estimator, a fairness-aware GNN with adversarial debiasing, and a precise unlearning mechanism built on Fisher Information Matrix analysis
Extensive experiments on real-world datasets show that FairGU clearly outperforms SOTA methods. 
It achieves large gains in fairness metrics while still maintaining strong accuracy and providing robust privacy protection against membership inference attacks.
Despite these strengths, FairGU's performance is sensitive to hyperparameters such as the selectivity factor and dampening strength, which may require careful tuning in practical applications. 
% Future work will focus on enhancing scalability and robustness, exploring adaptive hyperparameter optimization, and extending the framework to dynamic graphs and multi-sensitive attribute settings.

\bibliographystyle{ACM-Reference-Format}
\bibliography{acmart}

\appendix
\section{Ethical Use of Data and Informed Consent}
All datasets used in this study (e.g., Pokec-z, Pokec-n and Income) are publicly available under their respective research licenses. 
These datasets were released for research purposes and have been widely adopted in prior work on recommender systems and fairness-aware machine learning. 
They do not contain personally identifiable information beyond what has been made publicly accessible, and no additional data collection or annotation was conducted by the authors. 
As this work relies solely on secondary analysis of existing open datasets, no direct interaction with human participants occurred, and informed consent was not required. 
The study complies with the ethical use of data guidelines established by the research community.

\end{document}